\documentclass[journal]{IEEEtran}
\usepackage{times}  
\usepackage{helvet} 
\usepackage{courier}  
\usepackage[hyphens]{url}  
\usepackage{graphicx,subfigure} 
\usepackage{graphicx}  
\usepackage{epsfig}
\usepackage{amsmath}
\usepackage{amssymb}
\usepackage{epstopdf}
\usepackage{multirow}
\usepackage{setspace}
\usepackage{enumerate}
\usepackage{color}
\newcommand{\bm}[1]{\mbox{\boldmath{$#1$}}}
\newcommand{\ie}{{\em i.e.}}           

\makeatletter
\def\hlinew#1{%
  \noalign{\ifnum0=`}\fi\hrule \@height #1 \futurelet
   \reserved@a\@xhline}

\ifCLASSINFOpdf
\else
\fi

\begin{document}

\title{Class Distribution Alignment for Adversarial Domain Adaptation}

\author{Wanqi~Yang, Tong~Ling, Chengmei~Yang, Lei~Wang, Yinghuan~Shi, Luping~Zhou, Ming~Yang
\thanks{Wanqi Yang, Tong Ling, Chengmei Yang and Ming Yang are with the School of Computer Science and Technology, Nanjing Normal University, Nanjing, 210046, China. \protect (e-mail: yangwq@njnu.edu.cn,myang@njnu.edu.cn).}
\thanks{Yinghuan Shi is with the State Key Laboratory for Novel Software Technology, Nanjing University, Nanjing,
        210046, China. \protect (e-mail: syh@nju.edu.cn).}
\thanks{Lei Wang is with the School of Computing and Information Technology, University of Wollongong, Australia. \protect (e-mail: leiw@uow.edu.au).}
\thanks{Luping Zhou is with the School of Electrical and Information Engineering at the University of Sydney, Australia. \protect (e-mail: luping.zhou@sydney.edu.au).}
        }


\maketitle

\begin{abstract}
Most existing unsupervised domain adaptation methods mainly focused on aligning the marginal distributions of samples between the source and target domains. This setting does not sufficiently consider the class distribution information between the two domains, which could adversely affect the reduction of domain gap. To address this issue, we propose a novel approach called Conditional ADversarial Image Translation (CADIT) to explicitly align the class distributions given samples between the two domains. It integrates a discriminative structure-preserving loss and a joint adversarial generation loss. The former effectively prevents undesired label-flipping during the whole process of image translation, while the latter maintains the joint distribution alignment of images and labels. Furthermore, our approach enforces the classification consistence of target domain images before and after adaptation to aid the classifier training in both domains. Extensive experiments were conducted on multiple benchmark datasets including \emph{Digits}, \emph{Faces}, \emph{Scenes} and \emph{Office31}, showing that our approach achieved superior classification in the target domain when compared to the state-of-the-art methods. Also, both qualitative and quantitative results well supported our motivation that aligning the class distributions can indeed improve domain adaptation. 
\end{abstract}


\IEEEpeerreviewmaketitle

\section{Introduction}
The excellent performance of deep learning crucially depends on the availability of a large amount of labeled data. However, this requirement cannot be well met in many real scenarios, and the absence of labels in a target domain makes classification difficult. Leveraging labeled data from auxiliary source domains has been shown as a promising way to improve this situation. Nevertheless, the inevitable discrepancy of data distributions between the source and target domains poses new challenges. This has made unsupervised domain adaptation (UDA) an active research topic in artificial intelligence in the recent years.

Previous deep domain adaptation methods \cite{long2015learning,long2016unsupervised,long2017deep} learn transferable feature representation across different domains. It embeds domain adaptation modules into several high layers of a CNN network to narrow the distribution gap. Recently, inspired by the powerful generative adversarial network (GAN) \cite{goodfellow2014generative}, a few adversarial domain adaptation methods have been developed \cite{tzeng2017adversarial,long2018conditional,chen2019joint}. They leveraged an adversarial setting to learn a domain discriminator (to distinguish from which domain a latent feature representation was derived) and domain-invariant feature generator(s) (to confuse the discriminator). However, these methods usually transform features rather than images, which prevent them from producing interpretable and visualizable transformation that is often highly desirable in practical applications.


\begin{figure}
    \centering
    \includegraphics[width=0.96\columnwidth]{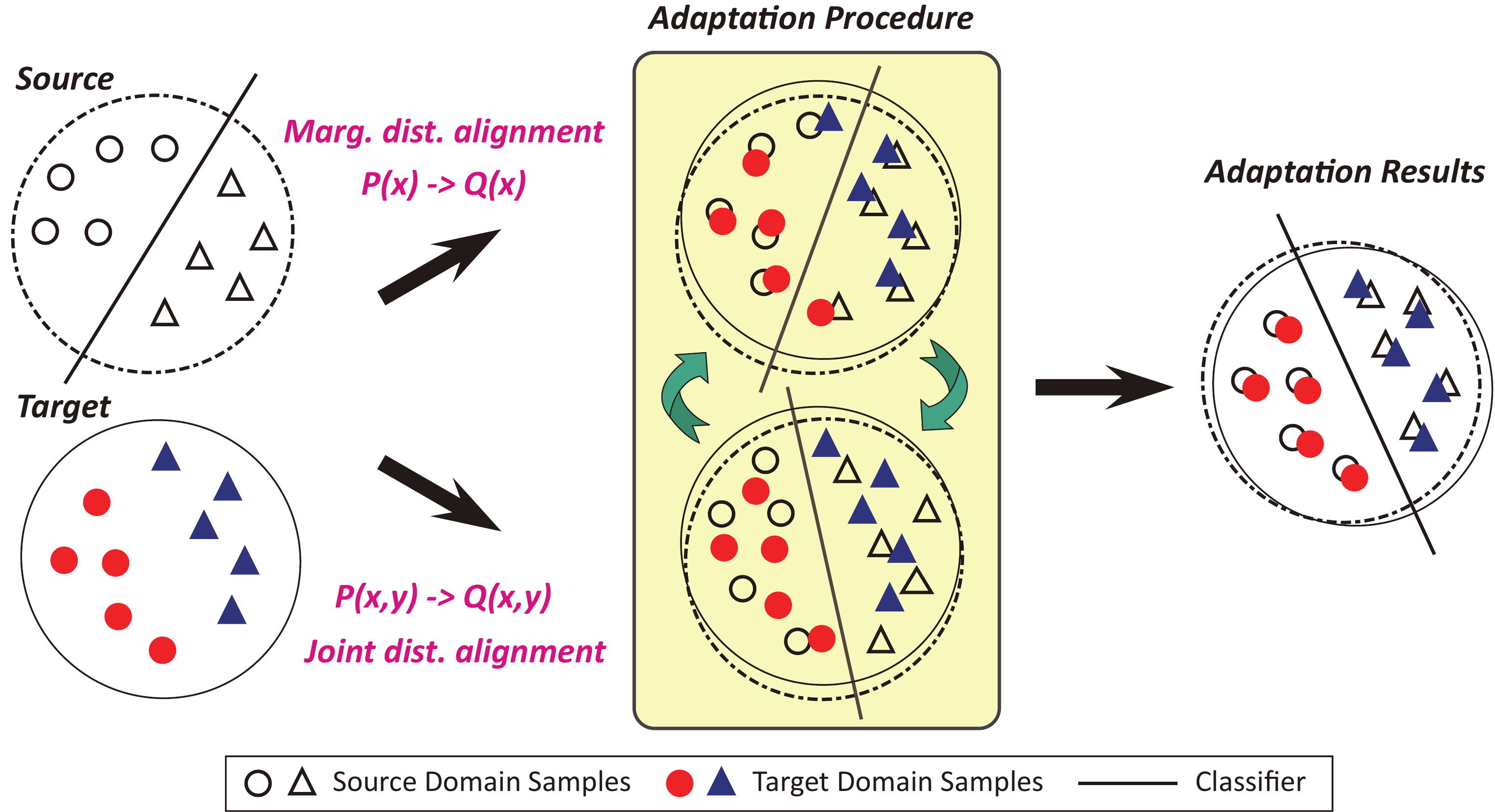}
    \caption{\small{An illustration of the domain adaptation in our method. In the light yellow box, \textbf{top}: only aligning the marginal distributions could make some target samples are misclassified; \textbf{bottom}: only aligning the joint distributions might cause the undesired alignment of samples. Considering both of them, our method will be able to better align the class distributions given samples (\emph{i.e.,} $P(y|x)$ and $Q(y|x)$), and thus helps to improve classification in the target domain.}}\label{example}
\end{figure}

This paper follows a special yet important setting in adversarial domain adaption: adversarial image translation between different domains, which leverages GAN to generate source-like or target-like images for domain adaptation. This approach can effectively deal with the changes in appearance or texture between domains, \ie, style transfer. Several methods along this line \cite{bousmalis2017unsupervised,sankaranarayanan2018generate,russo2018source,hoffman2018cycada,fu2019geometry} have been proposed in the recent years, well demonstrating their efficacy. Usually, these methods take a fundamental assumption that \emph{when the images from different domains conform a same marginal distribution after domain adaptation, their joint distribution with labels can be expected to be aligned at the same time}. Nevertheless, this assumption could be too strong to hold in many cross-domain scenarios. Once this happens, these methods will suffer from ONLY aligning the marginal distributions and deliver less satisfactory classification in the target domain. This case is illustrated in the top of the light yellow box in Fig. \ref{example} to better present the potential issue. As shown, if only aligning the marginal distributions, some target samples could still be misclassified because the class-specific samples between the two domains have not been well aligned, that is, the class distributions between the two domains are not sufficiently considered.

\begin{table}
\begin{center}
\renewcommand{\arraystretch}{1}
\setlength{\tabcolsep}{5pt}
\caption{\small{Difference in existing domain adaptation methods.}}\label{difference}
\begin{tabular}{l|c|c}\hline
\textbf{Methods}&$P(x)\rightarrow Q(x)$&$P(x,y)\rightarrow Q(x,y)$\\\hline\hline
ADGAN \cite{sankaranarayanan2018generate}&\checkmark&\\
DupGAN \cite{hu2018duplex}&\checkmark&\\
SBADA \cite{russo2018source}&\checkmark&\\
ACAL \cite{ehsan2019augmented}&\checkmark&\\
JAN \cite{long2017deep}&&\checkmark\\
CDAN \cite{long2018conditional}&&\checkmark\\
CADIT (ours)&\checkmark&\checkmark\\\hline
\end{tabular}
\end{center}
\end{table}

To address the above issue, \emph{\textbf{we explicitly propose to align the class probability distributions given samples between the source and target domains}}\footnote{For brevity, ``the class probability distribution given samples'' is abbreviated as ``the class distribution''.}, instead of merely aligning the marginal distributions. This is because the class distribution crucially determines the classification result of a target sample. Also, it was proved by the \emph{Theorem 1} in the literature \cite{ben2010theory} that the expected target error is upper-bounded by three terms: 1) source error, 2) domain divergence, and 3) the difference in labeling functions across domains. The difference of labeling function can be significantly affected by the difference of the class distributions between the two domains, thus we do our best to align the class distributions to reduce the target error. Note that according to the Bayes' rule, when both marginal distribution $P(x)$ and joint distribution $P(x,y)$ are aligned, the class distribution $P(y|x)$ shall also be aligned. Therefore, we propose a novel Conditional ADversarial Image Translation approach (CADIT) to simultaneously align (1) the marginal distributions and (2) the joint distributions between the two domains (see Fig. \ref{example}). Only aligning one of them will not ensure the desired alignment of the class distributions. Table \ref{difference} lists the main difference of existing domain adaptation methods from our method. It is worth noting that a few methods \cite{long2017deep,long2018conditional} have attempted to approximately align the joint distributions. However, due to the unavailability of target sample labels, effectively aligning the joint distributions still remains an open issue (see the \textbf{Remark} for detailed differences).

Specifically, to align the marginal distributions, we adopt the widely-used CycleGAN \cite{zhu2017unpaired} as a base network of adversarial image translation. Nevertheless, we found that CycleGAN, as an unsupervised image translation method, does not necessarily maintain the class labels of samples during translation, which is however important in our method. To deal with this issue, we put forward a discriminative structure-preserving (\emph{DSP}) loss to prevent the undesired flips of class labels. Also, to align the joint distributions, we develop a novel joint adversarial generation (\emph{JAG}) loss in both domains to jointly distinguish the authenticity/fakeness of the samples and their labels in an adversarial way. Since the labels of target samples are not available, we obtain their pseudo-labels and utilize the entropy-based prediction confidence to handle the uncertainty of these pseudo-labels. The target samples with more reliable pseudo-labels are allowed to contribute more to the joint distribution alignment.
Besides, we incorporate a classifier into each discriminator in both domains to simultaneously perform the image generation and classification. Considering that the two classifiers could be affected by the image generation and thus produce inconsistent classification predictions for a same target sample, we impose a classification consistence constraint (\emph{CLC}) loss on their predictions between the two domains to address this issue. By integrating the above components, we produce a more efficient and more interpretable unsupervised domain adaptation approach. In sum, the main contributions of this paper include:
\begin{itemize}
\item We propose a novel adversarial domain adaptation approach, CADIT, to explicitly strive to align the class distributions between the two domains.
\item To realize the proposed approach, we designed and effectively integrated three novel losses: \emph{DSP} loss, \emph{JAG} loss and \emph{CLC} loss to impose the key requirements.
\item Both qualitative and quantitative experimental results on multiple benchmark datasets (\emph{i.e.,} \emph{Digits}, \emph{Faces}, \emph{Scenes} and \emph{Office31}) validate the effectiveness and advantage of the proposed CADIT with respect to the state-of-the-art methods.
\end{itemize}

\section{Related Work}
Recently, unsupervised domain adaptation has aroused a great interest of researchers. Existing methods can be roughly classified into two categories as follows.

\textbf{Deep Domain Adaptation Methods.}
This category of methods \cite{long2016unsupervised,long2017deep,pan2019transferrable} employed deep networks to learn the domain-invariant representation, which is transferable between different domains. Long \emph{et al.} \cite{long2015learning} introduced multiple domain adaptation modules into the high layers of deep convolutional network to match the mean embeddings of distributions according to a maximum mean discrepancy criterion. Afterwards, they \cite{long2017deep} proposed a joint maximum mean discrepancy criterion to align the distributions of multiple domain-specific fully connected layers. In \cite{roy2019unsupervised}, an unified deep domain adaptation framework with domain alignment layers was proposed to match the feature distributions between different domains.

\begin{figure*}
    \centering
    \includegraphics[width=2\columnwidth]{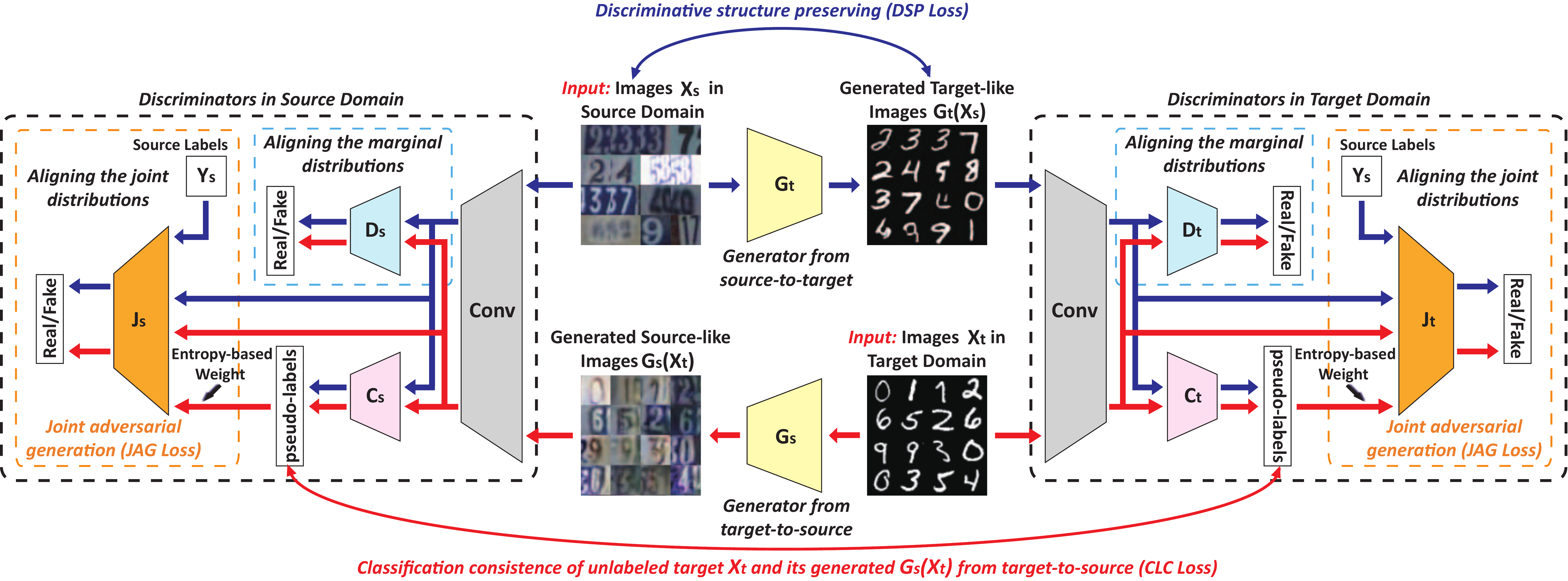}
    \caption{\small{The framework of our method. In the top middle part,  the images in source domain are translated by generator $G_t$ to target domain, while in the bottom middle part, the images in target domain are through generator $G_s$ to generate the source-like images. They together construct the cycle of cross-domain image translation. At the left and right ends, two domain-specific discriminators (shown in black dashed boxes) are developed to align both the marginal distributions and the joint distributions between the two domains and meanwhile predict the target samples. Each of the discriminators consists of a common convolutional module and three different output branches to respectively predict the real/fake samples, the classification labels, as well as the joint authenticity/fakeness of samples and labels.}}\label{framework}
\end{figure*}

\textbf{Adversarial Domain Adaptation Methods.}
This category of methods applied GANs to reduce the domain gap. There are several methods \cite{ganin2015unsupervised,tzeng2017adversarial,chen2019progressive} that developed the domain-invariant feature generators and a domain discriminator to distinguish their authenticity/fakeness. Tzeng \emph{et al.} \cite{tzeng2017adversarial} proposed a generalized framework to combine adversarial learning, discriminative feature learning and untied weight sharing. CDAN \cite{long2018conditional} built a multilinear map of the high-level feature representation and classification prediction between the source and target domains to adapt the map between them. To learn domain-invariant and semantic representations, a graph convolutional adversarial network \cite{ma2019gcan} was built to jointly perform the alignment of data structure, domain and class centroid.

Besides, several other methods \cite{sankaranarayanan2018generate,hu2018duplex,fu2019geometry} conducted adversarial image translation for domain adaptation by generating the source-like or target-like images from real images. Coupled GAN (CoGAN) \cite{liu2016coupled} trained a coupled generative model to learn the joint data distributions of the two domains. UNIT \cite{liu2017unsupervised} extended CoGAN to learn image translation by integrating variational autoencoder and GAN. ADGAN \cite{sankaranarayanan2018generate} simultaneously performed image translation and learned the shared feature embedding, by imposing the source and target distributions to be close in a latent space. SBADA \cite{russo2018source} employed the structure of CycleGAN for image translation and proposed target self-labeling and class consistence between source domain images and their circle generation results.

\textbf{Remark}.
Note that, our method belongs to adversarial image translation for domain adaptation that has a compelling visualization and interpretation of generation results. For the above existing methods (see Table \ref{difference}), these methods ADGAN \cite{sankaranarayanan2018generate}, DupGAN \cite{hu2018duplex}, SBADA\cite{russo2018source}, ACAL \cite{ehsan2019augmented} only aligned the marginal distributions of samples between the two domains and this, as we have argued, may not be sufficient for reducing domain gap. On the other hand, the two methods \cite{long2017deep,long2018conditional} attempted to approximately align the joint distributions by aligning the high-level feature representation and the prediction between two domains. However, such an alignment could be affected by the incorrect predictions of target samples, which is effectively mitigated in our method by considering the prediction uncertainty. Also, these two methods did not consider the alignment of the marginal distributions and this will not help to pursue the class distribution alignment. They will be experimentally compared with our method (see Table \ref{facescenerst}). 
Very recently, several works have began to concern the class-level alignment in domain adaption from different aspects \cite{ma2019gcan,chen2019joint,chen2019progressive,wen2019bayesian}. This supports the timeliness and significance of our work from another perspective. At the same time, we highlight that having key differences from our work, these methods are not designed for cross-domain image translation, and do not explicitly strive to align the class distributions as our method either.

\section{Our Approach}
Formally, given a source domain $\mathcal{D}_s=\{x_s^i,y_s^i\}_{i=1}^{n_s}$ and a target domain $\mathcal{D}_t=\{x_t^i\}_{i=1}^{n_t}$ (without known labels), $x_s$ and $x_t$ represent the images in the source and target domains, respectively, and $y_s$ represents the label of the source domain. In the target domain, the label $y_t$ is not available. The sample $x_s$ (or $x_t$) obeys the marginal distribution $P(x_s)$ (or $Q(x_t)$). Its joint distribution with labels and class distribution are denoted as $P(x_s,y_s)$ and $P(y_s|x_s)$ (or $Q(x_t,y_t)$ and $Q(y_t|x_t)$), respectively. Due to the discrepancy of the distributions between the two domains, the classifier trained on the source domain cannot be directly applied to the target domain. Thus, the task of UDA is to reduce the domain discrepancy to improve the generalization of the classifier on target samples. To achieve this, existing methods usually align either marginal or joint distributions by various strategies. Differently, we propose to align both of them in order to make the corresponding class distributions well matched, and this is implemented by conducting GAN-based image translation between the two domains.

Specifically, we design two GANs $(G_t, D_t)$ and $(G_s, D_s)$ to generate a target-like sample $G_t(x_s)$ from the source sample $x_s$ and a source-like sample $G_s(x_t)$ from the target sample $x_t$. For simplified notation,  $G_t(x_s)$ and $G_s(x_t)$ are abbreviated as $\hat{x}_{st}$ and $\hat{x}_{ts}$, respectively. Then, the image translation between the two domains is realized by:
\begin{enumerate}[(1)]
\item \textbf{(marginal distribution alignment)} making the marginal distribution $P(\hat{x}_{st})$ (translated from the source domain) align to $Q(x_t)$ (in the target domain), and $Q(\hat{x}_{ts})$ (translated from the target domain) align to $P(x_s)$ (in the source domain), as well as ensuring $\hat{x}_{st}$ to be well associated with its own source label $y_s$;

\item \textbf{(joint distribution alignment)} making the joint distribution $P(\hat{x}_{st},y_s)$ (translated from the source domain) align to the joint distribution $Q(x_t,y_t)$ (in the target domain), and $Q(\hat{x}_{ts},y_t)$ (translated from the target domain) align to $P(x_s,y_s)$ (in the source domain).
\end{enumerate}


By enforcing the two alignments in steps (1) and (2), we can expect the corresponding class distributions $P(y_s|\hat{x}_{st})$ and $Q(y_t|x_t)$ (or $Q(y_t|\hat{x}_{ts})$ and $P(y_s|x_s)$) to be better matched after image translation. In turn, this will make the classifier trained by the labeled samples (translated from the source domain) more applicable to classify the unlabeled target samples.

The framework diagram of our method is depicted in Fig. \ref{framework} with three following parts: (1) a discriminative cycle-consistent GAN with two generators and two discriminators to align the marginal distributions between the two domains after image translation, (2) a joint adversarial generation (\emph{JAG}) loss in each discriminator to align the joint distributions of samples and their labels between the two domains, and (3) a discriminative structure-preserving (\emph{DSP}) loss and a classification consistence (\emph{CLC}) loss to prevent label flipping and enhance the classification performance. They are detailed as follows.

\subsection{Discriminative Cycle-consistent GAN}
To align the marginal distributions, we adopt widely-used CycleGAN \cite{zhu2017unpaired} as a base network to realize our method, due to its success on image-to-image translation. CycleGAN is used to build the two generators $G_t$ and $G_s$ to learn two source-to-target and target-to-source mapping respectively: $G_t:x_s\rightarrow x_t$ and $G_s:x_t\rightarrow x_s$, where a target-like image $\hat{x}_{st}$ (or a source-like image $\hat{x}_{ts}$) is generated to adapt to real images $x_t$ (or $x_s$). To reduce the feasible region of mapping functions, a cycle-consistent loss is used to make $G_s(G_t(x_{s}))\approx x_s$ and $G_t(G_s(x_{t}))\approx x_t$. Meanwhile, against $G_s$ and $G_t$, the two discriminators $D_s$ and $D_t$ are designed to distinguish the authenticity or fakeness of images in both the domains, respectively. Moreover, to simultaneously perform the image translation and classification, we additionally design two classifiers $C_s$ and $C_t$ based on their discriminators.

Accordingly, the adversarial generation loss in the target domain can be written as:
\begin{equation}\label{ganloss1}
\begin{split}
\mathcal{L}_{\text{DGAN}_t}(G_t,D_t,C_t)&=\mathbb{E}_{\hat{x}_{st}\sim P(\hat{x}_{st})}[\log(1-D_t(\hat{x}_{st}))]\\
&+\mathbb{E}_{x_t\sim Q(x_t)}[\log D_t(x_t)]\\
&+(1/n_s)\sum_i\mathcal{L}_{ce}(C_t(\hat{x}_{st}^i),y_s^i),\\
\end{split}
\end{equation}
where $\mathcal{L}_{ce}(\cdot)$ represents the cross-entropy loss to train the classifier $C_t$ in the target domain by using the samples $\hat{x}_{st}$ translated from source-to-target.

Likewise, in the source domain, the classifier $C_s$ is trained by the real source samples $x_s$. Also, the adversarial generation loss in the source domain can be written as:
\begin{equation}\label{ganloss2}
\begin{split}
\mathcal{L}_{\text{DGAN}_s}(G_s,D_s,C_s)&=\mathbb{E}_{\hat{x}_{ts}\sim Q(\hat{x}_{ts})}[\log(1-D_s(\hat{x}_{ts}))]\\
&+\mathbb{E}_{x_s\sim P(x_s)}[\log D_s(x_s)]\\
&+(1/n_s)\sum_i\mathcal{L}_{ce}(C_s(x_s^i),y_s^i).\\
\end{split}
\end{equation}

Also, as in CycleGAN, the cycle-consistent loss is calculated as:
\begin{equation}\label{cyc}
\begin{split}
\mathcal{L}_{cyc}(G_s,G_t) &= \mathbb{E}_{x_s\sim P(x_s)}[||G_s(G_t(x_s))-x_s||_1]\\
&+\mathbb{E}_{x_t\sim Q(x_t)}[||G_t(G_s(x_t))-x_t||_1].\\
\end{split}
\end{equation}

\emph{\textbf{DSP Loss}}.
Since CycleGAN is originally designed for unsupervised image translation, it does not guarantee to maintain the label of images during image translation, which is however important for our alignment. As a result, an image $\hat{x}_{st}$ (translated from source-to-target) could be misaligned with its source domain label $y_s$. To prevent the label flipping (ensuring the label consistence before and after image translation), we propose a discriminative structure-preserving (\emph{DSP}) loss to make the generated images (from source-to-target) preserve their discriminative class structure in the source domain:
\begin{equation}\label{DSPloss}
\mathcal{L}_{\text{DSP}}(G_t)=\sum_{i\neq j}\bm{d}(\hat{x}_{st}^i,\hat{x}_{st}^j)*W_{ij},
\end{equation}
where $\bm{d}(\hat{x}_{st}^i,\hat{x}_{st}^j)$ represents the distance between any two generated images $\hat{x}_{st}^i$ and $\hat{x}_{st}^j$ (from source-to-target), calculated by Euclidean distance of their features. $W_{ij}$ is the weight between them (1 for the same class, -1 for different classes), according to their source domain labels. In this way, the generated images are encouraged to be similar when they belong to a same class and dissimilar otherwise. Thus, a generated image is subject to other generated images during image translation, so its label is difficult to be flipped, which has also been validated in the experiments (see Fig. \ref{digitvisualizationsvhn} and Tables \ref{ablationstudy1}-\ref{ablationstudy2}).

\emph{\textbf{CLC Loss}}.
As shown above, during domain adaptation, the two classifiers $C_t$ and $C_s$ are used to predict the labels of target sample $x_t$ and its generated sample $\hat{x}_{ts}$ (from target-to-source), respectively. Since the two classifiers could be affected by the image translation and thus produce inconsistent classification predictions, we impose a classification consistence constraint (\emph{CLC}) loss on their predictions to aid the classifier training as:
\begin{equation}\label{cons}
\begin{split}
\mathcal{L}_{\text{CLC}}(C_s,C_t,G_s) &= \mathbb{E}_{x_t\sim Q(x_t)}[||C_t(x_t)-C_s(\hat{x}_{ts})||_1].\\
\end{split}
\end{equation}
Obviously, the \emph{CLC} loss is also useful to prevent the label flipping during image transaltion. Note that since the predictions of source samples in the two classifiers are constrained by their labels in Eqs.(\ref{ganloss1})(\ref{ganloss2}), their consistence constraint is safely omitted.

\subsection{Joint Adversarial Generation}
To align the joint distributions, based on CycleGAN, we develop two novel joint discriminators $J_t$ and $J_s$ in the two domains to jointly distinguish the authenticity/fakeness of samples and labels.

\emph{\textbf{JAG Loss}}.
Note that the true label $y_t$ of target sample $x_t$ is not available during training, so we adopt the classification prediction $C_t(x_t)$ as its pseudo-label. However, the pseudo-labels may be incorrect that could affect the alignment of the joint distributions. To alleviate the impact of false pseudo-labeled samples, we expect to set higher confidence weight for more reliable pseudo-labels that are encouraged to contribute more to the joint distribution alignment. Inspired by entropy minimization to handle the uncertainty, we design an entropy-based confidence weight $\gamma(x_t)=1-H(x_t)/\log C$ for $x_t$ and its pseudo-label, where $H(x_t)$ is the entropy of its prediction probabilities on all classes, and $C$ is the number of classes.

In the joint discriminator $J_t$ of the target domain, we expect to align the joint distributions of real and generated images (from source-to-target) associated with their respective labels, and thus design a joint adversarial generation (\emph{JAG}) loss in the target domain as:
\begin{equation}\label{jointganloss1}
\begin{split}
\mathcal{L}_{\text{JGAN}_t}&(G_t,J_t,C_t)=\mathbb{E}_{(\hat{x}_{st},y_s)\sim P(\hat{x}_{st},y_s)}[\log(1-J_t(\hat{x}_{st},y_s))]\\
&+\mathbb{E}_{(x_t,y_t)\sim Q(x_t,y_t)}[\gamma(x_t)\log J_t(x_t,C_t(x_t))],\\
\end{split}
\end{equation}
where the paired samples and labels $(\hat{x}_{st},y_s)$ or $(x_t,C_t(x_t))$ are simultaneously fed into the joint discriminator $J_t$ to distinguish the authenticity/fakeness. The lower the weight $\gamma(x_t)$, the less the target sample $x_t$ works. On the contrary, those target samples with higher weight are more useful to align the joint distributions.

Likewise, in the joint discriminator $J_s$ of the source domain,
we also introduce the entropy-based confidence weight $\gamma(\hat{x}_{ts})$ to balance its possibly-incorrect pseudo-label. Thus, the \emph{JAG} loss in the source domain can be written as:
\begin{equation}\label{jointganloss2}
\begin{split}
&\mathcal{L}_{\text{JGAN}_s}(G_s,J_s,C_s)=\mathbb{E}_{(x_s,y_s)\sim P(x_s,y_s)}[\log J_s(x_s,y_s)]\\
&\ \ +\mathbb{E}_{(\hat{x}_{ts},y_t)\sim Q(\hat{x}_{ts},y_t)}[\gamma(\hat{x}_{ts})\log(1-J_s(\hat{x}_{ts},C_s(\hat{x}_{ts})))].\\
\end{split}
\end{equation}

\begin{figure*}
\centering
\includegraphics[width=1.8\columnwidth]{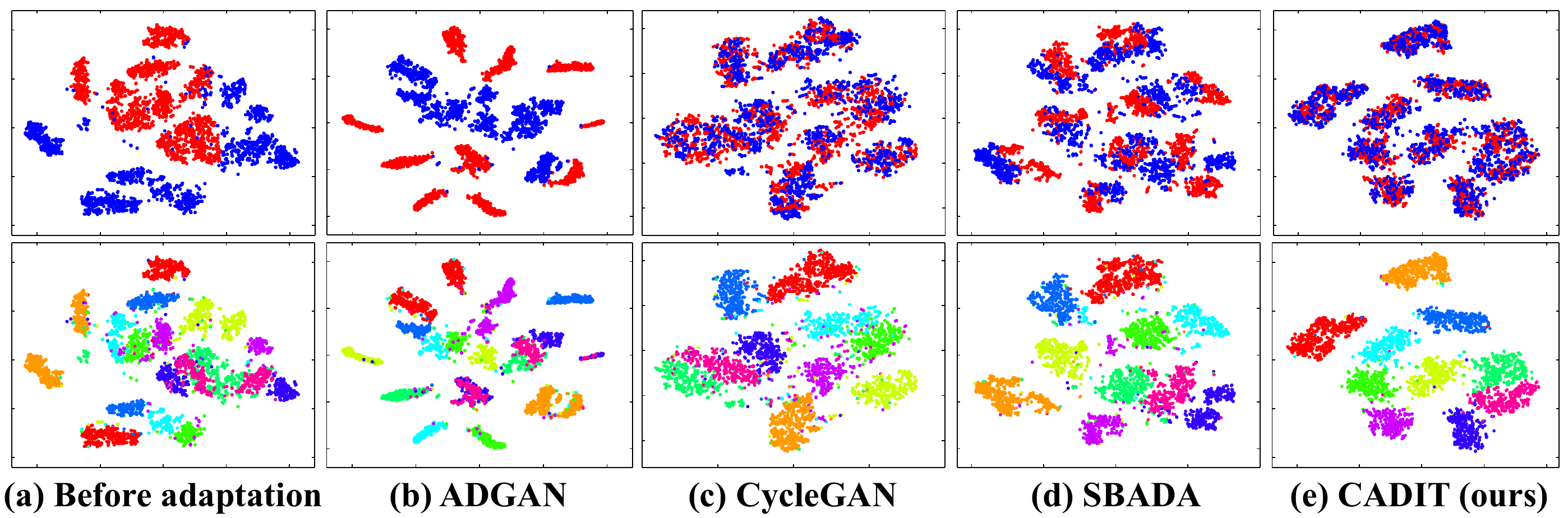}
\caption{\small{Visualization of the $t$-SNE results before and after M$\rightarrow$U adaptation with four different methods. \textbf{Top}: each sub-figure colorizes different domains (MNIST in red and USPS in blue). \textbf{Bottom}: each sub-figure colorizes at the level of ten digit classes (0-9).}}\label{tSNE_mtou}
\end{figure*}

\subsection{Main Objective}
In summary, to effectively align the class distributions for domain adaptation, we integrate the aforementioned seven terms in Eqs. (\ref{ganloss1})-(\ref{jointganloss2}) into a whole objective function as below:
\begin{equation}\label{objective}
\begin{split}
&\mathcal{L}(G_s,G_t,C_s,C_t,D_s,D_t,J_s,J_t)\\
=&\mathcal{L}_{\text{DGAN}_t}(G_t,D_t,C_t)+\mathcal{L}_{\text{DGAN}_s}(G_s,D_s,C_s)\\
+&\mathcal{L}_{\text{JGAN}_t}(G_t,J_t,C_t)+\mathcal{L}_{\text{JGAN}_s}(G_s,J_s,C_s)\\
+&\lambda_1\mathcal{L}_{\text{cyc}}(G_s,G_t)+\lambda_2\mathcal{L}_{\text{CLC}}(C_s,C_t,G_s)+\lambda_3\mathcal{L}_{\text{DSP}}(G_t),\\
\end{split}
\end{equation}
where parameters $\lambda_1,\lambda_2,\lambda_3$ are the weight of cycle-consistent term (for image translation), \emph{CLC} loss term (for target samples translated from target-to-source), and \emph{DSP} loss term (for source samples translated from source-to-target), respectively. For the values of $\lambda_1,\lambda_2,\lambda_3$ in the experiments, we empirically set them as ($5, 0.1, 10^{\text{-}4}$) for \emph{Digits}, ($10, 0.5, 10^{\text{-}3}$) for \emph{Faces} and \emph{Scenes}, and ($10, 10^{\text{-}3}, 10^{\text{-}4}$) for \emph{Office31} respectively. Also, we found that our method is robust against the variation of three parameters.

Note that in Eq. (\ref{objective}), two generators $G_s,G_t$ and two classifiers $C_s,C_t$ are solved by minimizing the above objective, while four discriminators $D_s,J_s,D_t$ and $J_t$ are solved by maximizing the above objective. As the traditional GAN, the generators and discriminators are trained in an adversarial manner, and they are optimized alternately. In the training phase, the labeled source samples and unlabeled target samples are used to train our model. In the testing phase, we adopt the learned classifier $C_t$ to predict the labels of the unlabeled target samples.

\textbf{Network Architecture}.
We implement our method CADIT using PyTorch on a single NVIDIA GeForce GTX 2080TI GPU. For the generators $G_s$ and $G_t$, we adopt two different network structures on all the datasets due to their different image sizes. For \emph{Digits}, we construct them by designing a simple two-layer convolutions and two-layer deconvolutions. For the other datasets, we apply the same architecture of the generators as that in CycleGAN, which contains two stride-2 convolutions, six residual blocks, and two fractionally-strided convolutions with stride 1/2. For the discriminators, we adopt a same network structure in all the datasets, which includes a common part and three different branches, as shown in Fig. \ref{framework}. The common part is implemented by two stride-2 convolutions and a fully-connection layer with the LeakyReLU activation, and all the three branches are a specific fully-connected layer. Also, for the adversarial generation loss function in $D_s, J_s, D_t, J_t$, we adopt mean squared error (MSE) loss for training.

\textbf{Network Training}.
We use the Adam solver \cite{kingma2015adam} with the related gradient weights of $0.5$ and $0.999$ to train our model. The base learning rate is initially set as $0.002$, and then decreases from $0.002$ to $0.0002$ as the number of iterations increases. The batch size is set as 64 for \emph{Digits}, 8 for \emph{Faces} and \emph{Scenes}, and 6 for \emph{Office31}, respectively. To reduce the training time on \emph{Faces} and \emph{Scenes}, we first train the CycleGAN network (by setting $\lambda_1$ as $10$) to initialize our generators, and then alternately optimize our generators and discriminators.

\section{Experiments}\label{experiments}
We evaluate the performance of our method on unsupervised domain adaptation for image classification. The experiments are conducted on three domain adaptation datasets: (1) \emph{Digits}, (2) \emph{Faces}, and (3) \emph{Scenes}. For all the experiments, we follow the standard unsupervised protocol: using the entire labeled samples in the source domain and unlabeled samples in the target domain. It is worth noting that the samples between the two domains are not paired. Also, the labels of samples in the target domain are not available during the training phase.

\begin{table}
 \renewcommand{\arraystretch}{1}
 \setlength{\tabcolsep}{10pt}
  \centering
  \caption{Comparison of the classification accuracy in M$\leftrightarrow$U adaptation. The results are in percent, and the best results are in bold.}\label{digitrecognition}
        \begin{tabular}{lccc}\hline
        \textbf{Methods}&\textbf{M$\rightarrow$U}&\textbf{U$\rightarrow$M}&\textbf{Average}\\\hline\hline
        \textbf{JDA} (2013)&67.28&59.65&63.47\\
        \textbf{TJM} (2014)&63.28&52.25&57.77\\
        \textbf{DANN} (2016)&91.11&74.01&82.56\\
        \textbf{DRCN} (2016)&91.80&73.67&82.74\\
        \textbf{CoGAN} (2016)&91.20&89.10&90.15\\
        \textbf{JGSA} (2017)&80.44&68.15&74.30\\
        \textbf{ADDA} (2017)&89.40&90.10&89.75\\
        \textbf{JAN} (2017)&93.61&93.15&93.38\\
        \textbf{CDAN} (2018)&93.27&94.25&93.76\\
        \textbf{RAAN} (2018)&89.00&92.10&90.55\\
        \textbf{MCD} (2018)&94.20&94.10&94.15\\
        \textbf{ADGAN} (2018)&92.80&90.80&91.80\\
        \textbf{CDRD} (2018)&95.05&94.35&94.70\\
        \textbf{SBADA} (2018)&95.41&93.49&94.45\\
        \textbf{TPN} (2019)&92.10&94.10&93.10\\
       \textbf{CADIT} (ours)&\textbf{96.74}&\textbf{95.04}&\textbf{95.89}\\\hline
        \end{tabular}
\end{table}

\subsection{Digits: MNIST$\leftrightarrow$USPS}
For \emph{Digits}, MNIST and USPS, two widely-used hand-written digit image datasets are viewed as two different domains. Following the same protocol in \cite{long2013transfer,long2014transfer}, we randomly sample 2,000 images in MNIST and 1,800 images in USPS to construct the source/target-domain dataset, and uniformly rescale all images to the size of 16$\times$16. We test two domain adaptation settings: (1) MNIST$\rightarrow$USPS (abbreviated as M$\rightarrow$U), and (2) USPS$\rightarrow$MNIST (abbreviated as U$\rightarrow$M). In each setting, we apply the learned classifier $C_t$ for digital image recognition (0-9) in the target domain.

We compare our method with fifteen representative UDA baselines: four traditional domain adaptation methods (JDA \cite{long2013transfer}, TJM \cite{long2014transfer}, JGSA \cite{zhang2017joint}, and TPN \cite{pan2019transferrable}), three deep domain adaptation methods (DANN \cite{ganin2016domain}, DRCN \cite{ghifary2016deep}, and JAN \cite{long2017deep}), four adversarial domain adaptation methods (ADDA \cite{tzeng2017adversarial}, CDAN \cite{long2018conditional}, RAAN \cite{chen2018re} and MCD \cite{saito2018maximum}), and four related adversarial image translation methods (CoGAN \cite{liu2016coupled}, ADGAN \cite{sankaranarayanan2018generate}, CDRD \cite{liu2018detach} and SBADA \cite{russo2018source}). Table \ref{digitrecognition} reports the average accuracy results,in which \emph{JAN, CDAN and SBADA were experimented by the publicly released codes, and the other twelve methods were reported by the literature.} For fair comparison, the baselines applied the same protocol of data as our method (using randomly sampled 2000/1800 images instead of the entire training sets). As observed, our method shows the best performance in both settings and the average. This verifies the effectiveness of our method.

\begin{figure*}
\centering
\includegraphics[width=1.6\columnwidth]{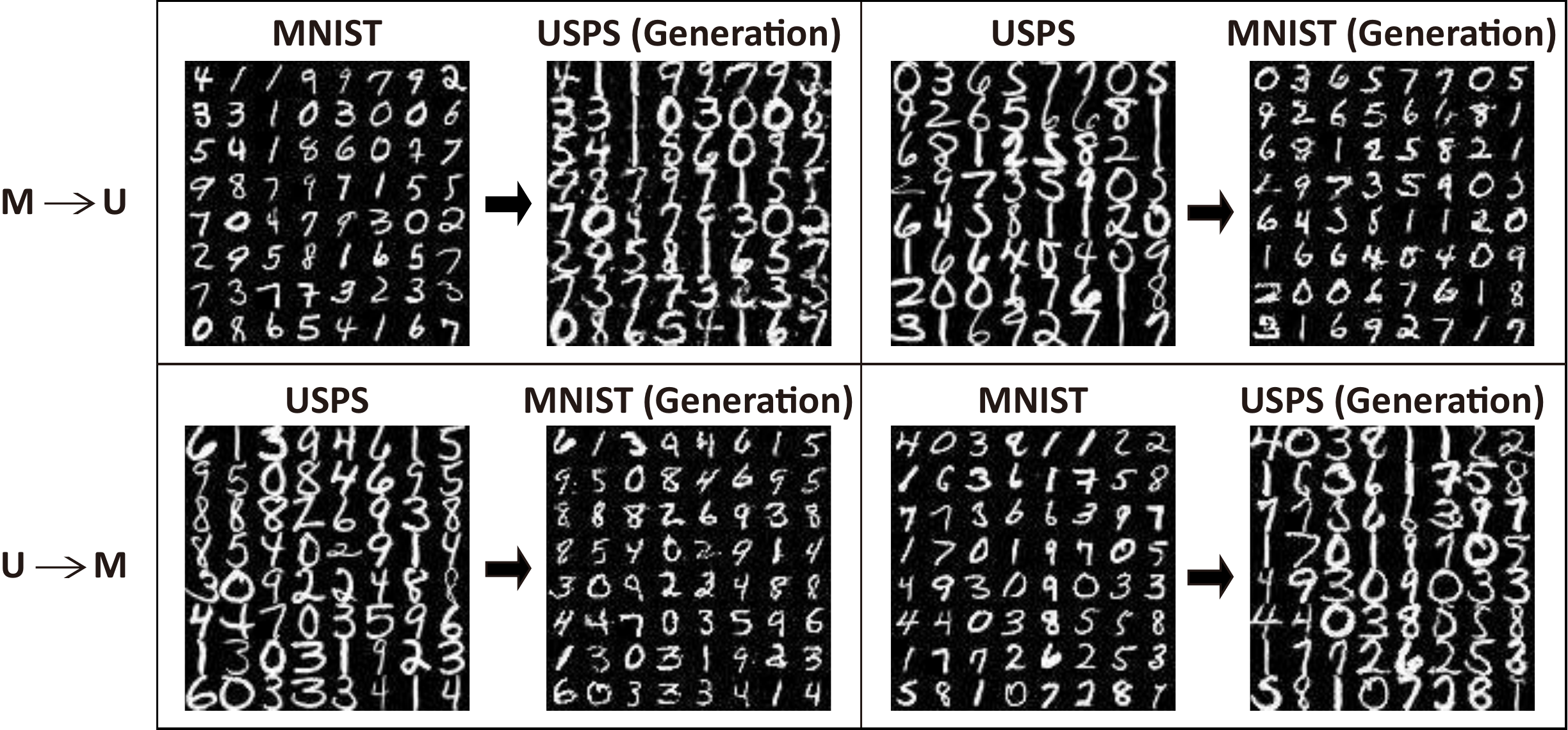}
\caption{\small{Examples of generated USPS-like images from MNIST and generated MNIST-like images from USPS during both M$\rightarrow$U adaptation (top) and U$\rightarrow$M adaptation (bottom) by using our CADIT.}}\label{digitvisualization}
\end{figure*}

Also, we show the $t$-SNE results of M$\rightarrow$U for different adversarial image translation methods in Fig. \ref{tSNE_mtou}. Specifically, we compare the real and generated images for CycleGAN, SBADA and CADIT, and compare the generated images from both domains for ADGAN. Due to their small size and grayscale intensity, we simply project their intensity values to 2D embedding space as same as SBADA \cite{russo2018source}. It is observed that the samples between the two domains are not aligned before image translation (see Fig. \ref{tSNE_mtou}(a)), while the samples after image translation via our method are best aligned and clustered than the other methods (see Fig. \ref{tSNE_mtou}(b)-(e)). Although the results of ADGAN show a good cluster structure, they are not well aligned between the two domains. This indicates that the class distributions between the two domains are well aligned in our method. In addition, we show the visualization results of typical examples in image translation between MNIST and USPS in Fig. \ref{digitvisualization}. As observed, the generated images can well capture the style of digital images in the current domain.
\begin{table}
 \renewcommand{\arraystretch}{1}
 \setlength{\tabcolsep}{10pt}
  \centering
  \caption{Classification accuracy for S$\leftrightarrow$M. The best results are in bold.}\label{svhnrst}
        \begin{tabular}{lccc}\hline
        \textbf{Methods}&\textbf{S$\rightarrow$M}&\textbf{M$\rightarrow$S}&\textbf{Average}\\\hline\hline
        \textbf{DANN} (2016)&73.90&35.70&54.80\\
        \textbf{DRCN} (2016)&81.97&40.05&61.05\\
        \textbf{ADDA} (2017)&76.00&-&-\\
        \textbf{JAN} (2017)&46.06&17.49&31.78\\
        \textbf{CDAN} (2018)&88.50&17.38&52.94\\
        \textbf{ADGAN} (2018)&84.70&36.40&60.55\\
        \textbf{SBADA} (2018)&76.10&61.10&68.60\\
        \textbf{CyCADA} (2018)&90.40&-&-\\
        \textbf{ACAL} (2019)&\textbf{93.90}&-&-\\
       \textbf{CADIT} (ours)&92.61&\textbf{61.47}&\textbf{77.04}\\\hline
        \end{tabular}
\end{table}

\subsection{Digits: SVHN$\leftrightarrow$MNIST}
\begin{figure*}
\centering
\includegraphics[width=1.6\columnwidth]{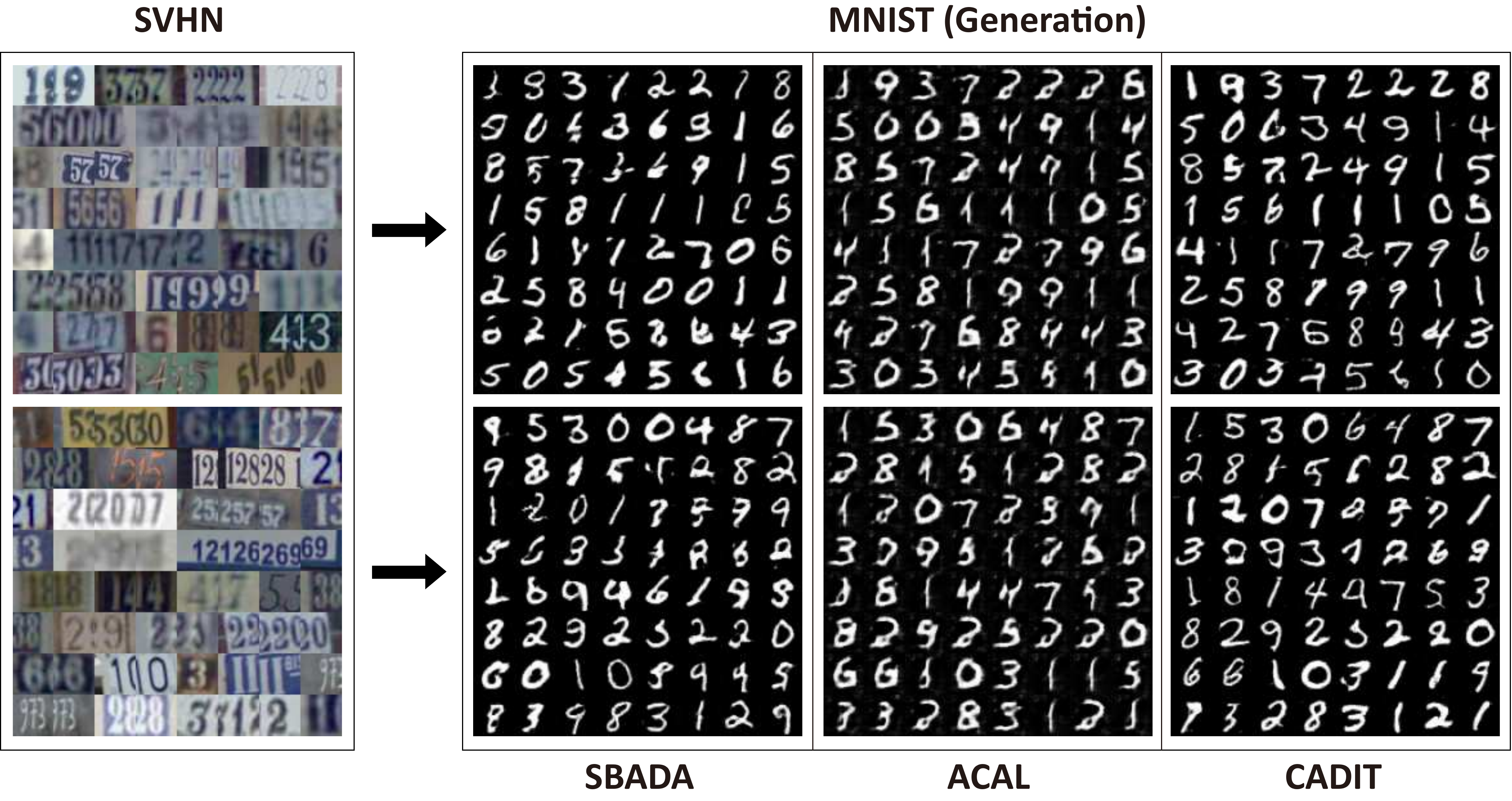}
\caption{\small{Examples of generated MNIST-like images from SVHN digital images by using three different methods (\emph{i.e.,} SBADA, ACAL and CADIT).}}\label{digitvisualizationsvhn}
\end{figure*}

SVHN \cite{netzer2011reading} was obtained by cropping house numbers in Google Street View images, and contains over 600k color images with a large variety of shapes and textures. Following the protocol \cite{tzeng2017adversarial}, we use the entire training data of SVHN and MNIST as two domains, and test two settings: (1) SVHN$\rightarrow$MNIST (abbreviated as S$\rightarrow$M), and (2) MNIST$\rightarrow$SVHN (abbreviated as M$\rightarrow$S), which presents a more challenging case of domain adaptation due to larger domain gap than M$\leftrightarrow$U. 

We compare our method with nine representative UDA baselines: 
five deep/adversarial domain adaptation methods (DANN \cite{ganin2016domain}, DRCN \cite{ghifary2016deep}, ADDA \cite{tzeng2017adversarial}, JAN \cite{long2017deep}, and CDAN \cite{long2018conditional}) and four latest and related adversarial image translation methods (ADGAN \cite{sankaranarayanan2018generate}, 
SBADA \cite{russo2018source}, CyCADA \cite{hoffman2018cycada}, and ACAL \cite{ehsan2019augmented}). For fair comparison, the baselines applied the same protocol of data with our method. We evaluate the performance of JAN and CDAN by their publicly released codes and meanwhile report the best results of the other seven methods in the literature in Table \ref{svhnrst}, compared with our method. As observed, the M$\rightarrow$S and average result of our method outperform the rest baselines. In addition, we show the visualization results of typical examples in image translation from SVHN to MNIST by using three different methods (SBADA, ACAL and CADIT) in Fig. \ref{digitvisualizationsvhn}. As observed, during image translation via SBADA, the labels of some digits are changed, \emph{e.g.,} `4' is often translated to '6', `1' is often translated to `9'. For ACAL, all the images of `2' are translated to `7'-like images, which make them difficult to distinguish. Also, all the images of `4' are not well generated. Compared with SBADA and ACAL, our CADIT can effectively prevent label flipping by using the proposed \emph{DSP} loss, and most of the generated digits seem more clear and better generated.

\subsection{Faces: Photo$\rightarrow$Sketch}
For \emph{Faces}, we use facial photo and sketch images as the source and target domains, respectively. Inspired by \cite{liu2018detach}, we construct a facial photo/sketch dataset by using the CelebA dataset \cite{liu2015deep} that contains more than 200K celebrity photos annotated with forty facial attributes. Specifically, we randomly select half of the CelebA dataset as the source domain, and then convert the other half into sketch images by using the method in \cite{huang2018multimodal} for the target domain. Since the number of attributes is large, we choose three common and representative attributes: ``glasses'', ``smiling'' and ``male'' (similar with that in \cite{liu2018detach}) to perform three individual binary classification tasks. These tasks are to recognize if a target sketch image has the corresponding attribute, respectively, following the standard UDA setting.
\begin{figure*}
    \centering
    \includegraphics[width=1.6\columnwidth]{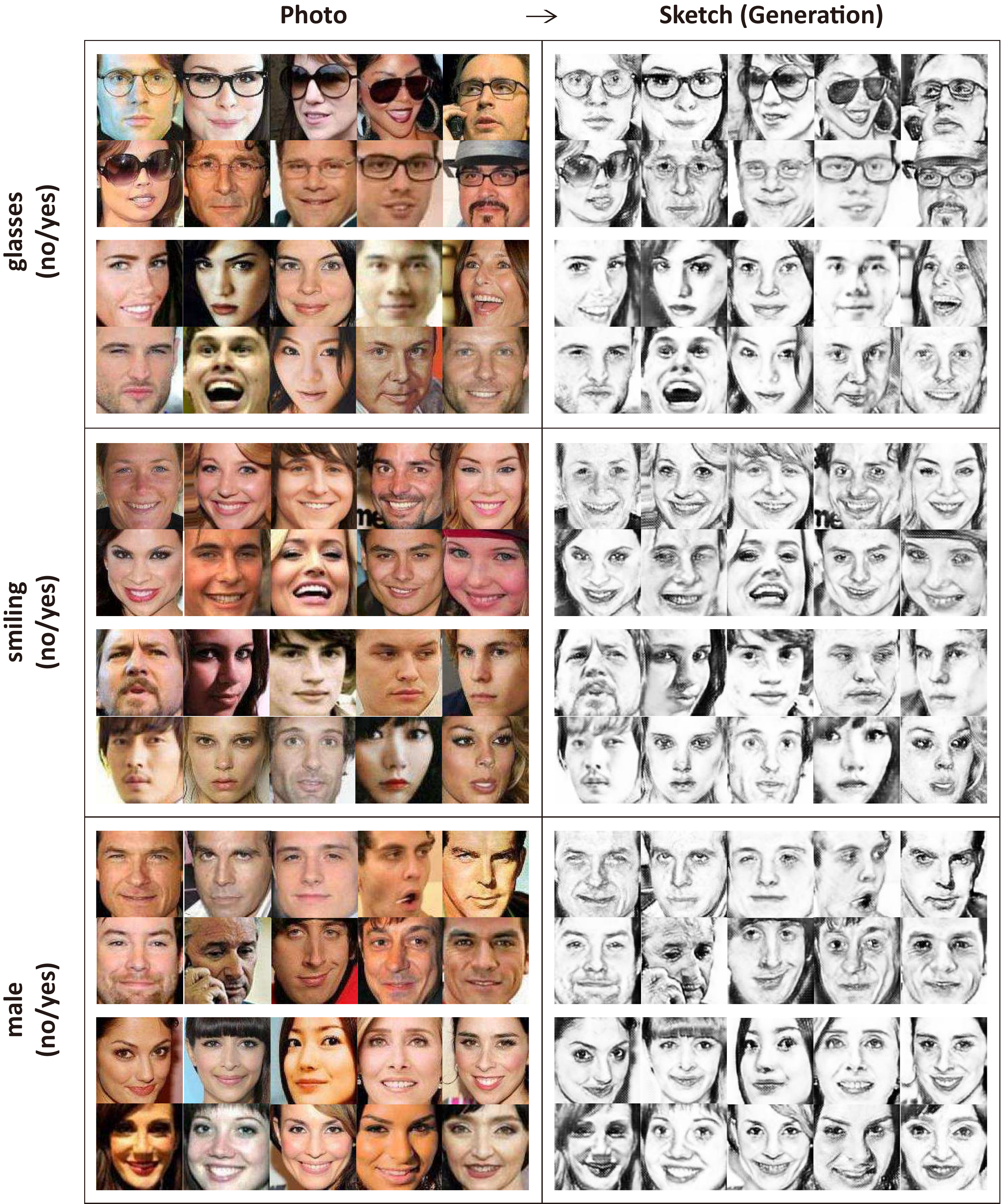}
    \caption{\small{Visualization of several typical examples of generated sketch images from facial photo images by using our CADIT.}}\label{facevisualization}
\end{figure*}

We compare our method with eight latest and most-related methods, JAN \cite{long2017deep}, CDAN \cite{long2018conditional},  ADGAN \cite{sankaranarayanan2018generate}, DupGAN \cite{hu2018duplex}, SBADA \cite{russo2018source}, DWT \cite{roy2019unsupervised}, ACAL \cite{ehsan2019augmented} and JDDA \cite{chen2019joint}. For fair comparison, we test these methods by utilizing their released source codes
and learning the best parameters. The average classification results of five independent runs on facial images are recorded in Table \ref{facescenerst1}. As observed, CADIT obtains the highest accuracy in all the compared methods for the three classification tasks, which shows that our method with jointly aligning the marginal and joint distributions is very effective for classifying target samples again. Also, the visualization results of typical examples in facial image translation from photo to sketch are illustrated in Fig. \ref{facevisualization}. The generated sketch-like images are clear and realistic, which shows that our method has a promising visualization performance during adaptation besides its superior classification performance.

\begin{table}
   \renewcommand{\arraystretch}{1}
   \setlength{\tabcolsep}{6pt}
   \centering
        \caption{Classification accuracy (mean $\pm$ std\%) on \emph{Faces}. The best results are in bold.}\label{facescenerst1}
        \begin{tabular}{lccc}\hline
        \multirow{2}{*}{\textbf{Methods}}&\multicolumn{3}{c}{\textbf{\emph{Faces}}}\\\cline{2-4}
        &\textbf{glasses}&\textbf{smiling}&\textbf{male}\\\hline\hline
        \textbf{JAN} (2017)&92.31$\pm$0.53&77.18$\pm$0.83&74.06$\pm$0.35\\
        \textbf{CDAN} (2018)&92.15$\pm$1.43&78.43$\pm$0.80&73.20$\pm$1.67\\
        \textbf{ADGAN} (2018)&92.43$\pm$1.71&79.66$\pm$1.73&73.59$\pm$0.85\\
        \textbf{DupGAN} (2018)&90.36$\pm$1.13&76.58$\pm$1.10&72.58$\pm$0.61\\
        \textbf{SBADA} (2018)&95.46$\pm$0.44&83.36$\pm$0.40&77.34$\pm$0.87\\
        \textbf{DWT} (2019)&93.83$\pm$0.51&82.68$\pm$0.49&78.79$\pm$0.58\\
        \textbf{ACAL} (2019)&94.90$\pm$0.64&83.11$\pm$0.58&76.48$\pm$0.41\\
        \textbf{JDDA} (2019)&93.96$\pm$0.63&82.59$\pm$0.91&75.37$\pm$0.92\\
        \textbf{CADIT} (ours)&\textbf{97.11}$\pm$0.20&\textbf{86.16}$\pm$0.65&\textbf{80.91}$\pm$0.45\\\hline
        \end{tabular}
\end{table}

\subsection{Scenes: Photo$\rightarrow$Painting}
For \emph{Scenes}, we use scene photo and painting images as the source and target domains, respectively. We collect 1,838 scene photo images from the publicly accessible dataset\footnote{https://people.eecs.berkeley.edu/\textasciitilde taesung\_park/CycleGAN/datasets/}, and then randomly select
half of the photo collection as the source domain. Afterwards, we construct painting images by applying the method in \cite{huang2018multimodal} on the rest half of the photo collection for the target domain. For performance evaluation, these scene images are manually labeled with three interest attributes: ``night'', ``winter'', and ``mountain''. Then, we test three binary classification tasks with UDA setting: using labeled source photo images to recognize target painting images with the corresponding attribute, respectively.

\begin{figure*}
    \centering
    \includegraphics[width=1.6\columnwidth]{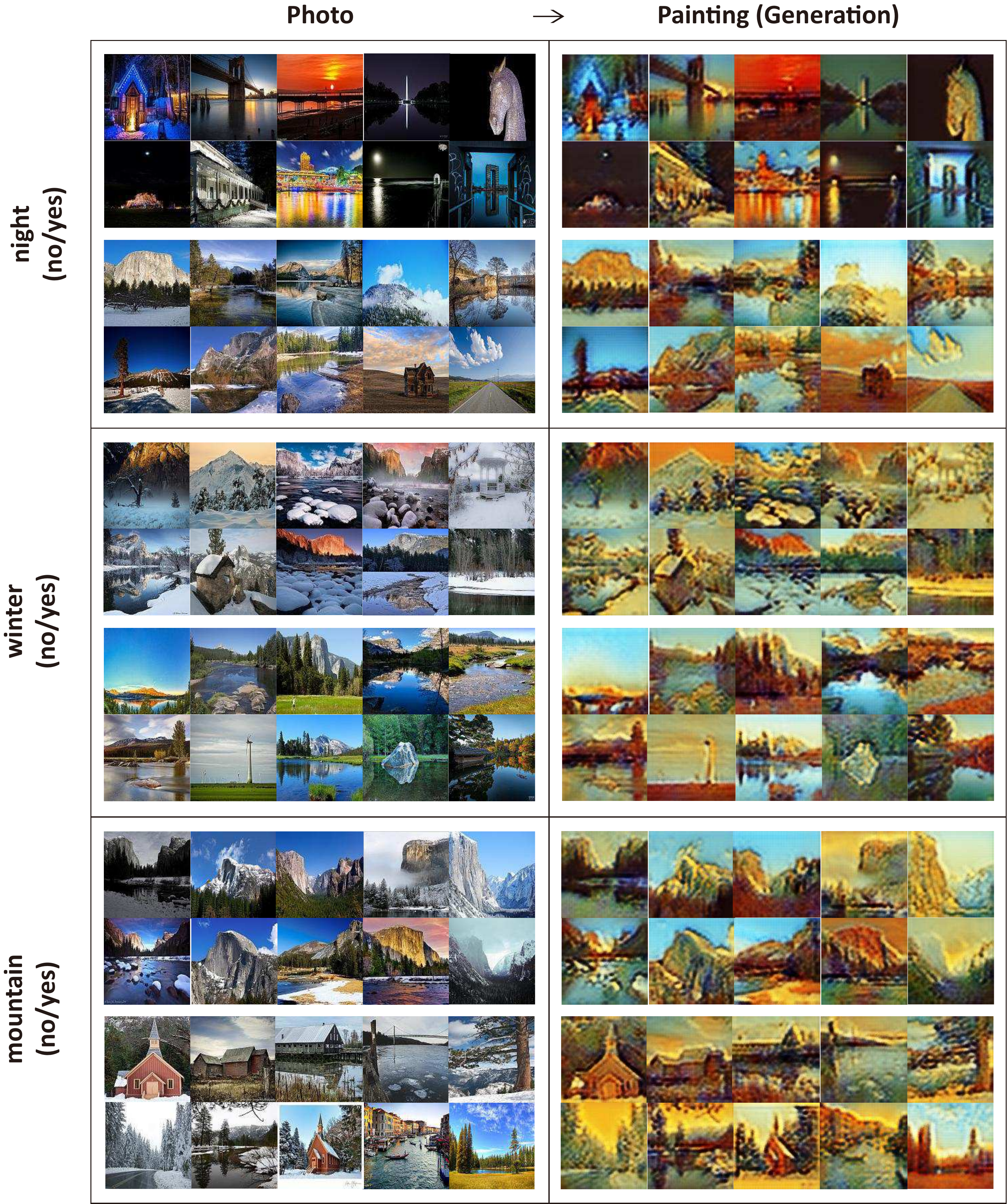}
    \caption{\small{Visualization of several typical examples of generated painting images from scene photo images by using our CADIT.}}\label{scenevisualization}
\end{figure*}

We compare CADIT with the eight related methods (same as \emph{Faces}) on \emph{Scenes}. The average classification results of five independent runs are recorded in Table \ref{facescenerst2}. It is observed that CADIT achieves the highest accuracy against the other methods for all the three classification tasks, which validates the efficacy of our method again. Also, Fig. \ref{scenevisualization} shows the visualization results of typical examples of image translation from photo to painting. Again, the clear and realistic image translation results are obtained by the proposed CADIT.
\begin{table}
   \renewcommand{\arraystretch}{1}
   \setlength{\tabcolsep}{6pt}
   \centering
        \caption{Classification accuracy (mean $\pm$ std\%) on \emph{Scenes}. The best results are in bold.}\label{facescenerst2}
        \begin{tabular}{lccc}\hline
        \multirow{2}{*}{\textbf{Methods}}&\multicolumn{3}{c}{\textbf{\emph{Scenes}}}\\\cline{2-4}
        &\textbf{night}&\textbf{winter}&\textbf{mountain}\\\hline\hline
        \textbf{JAN} (2017)&81.73$\pm$0.87&68.55$\pm$0.68&71.60$\pm$0.46\\
        \textbf{CDAN} (2018)&81.35$\pm$1.15&67.70$\pm$0.78&70.13$\pm$0.95\\
        \textbf{ADGAN} (2018)&81.83$\pm$0.15&68.16$\pm$0.68&71.03$\pm$0.23\\
        \textbf{DupGAN} (2018)&79.36$\pm$0.73&66.67$\pm$1.03&69.18$\pm$0.64\\
        \textbf{SBADA} (2018)&83.38$\pm$0.94&70.08$\pm$0.80&73.22$\pm$0.73\\
        \textbf{DWT} (2019)&82.39$\pm$0.74&68.92$\pm$0.31&72.02$\pm$0.27\\
        \textbf{ACAL} (2019)&83.56$\pm$0.48&69.72$\pm$0.79&72.50$\pm$0.78\\
        \textbf{JDDA} (2019)&84.11$\pm$0.42&71.83$\pm$0.91&74.73$\pm$0.91\\
        \textbf{CADIT} (ours)&\textbf{86.05}$\pm$0.61&\textbf{73.14}$\pm$0.63&\textbf{76.05}$\pm$0.58\\\hline
        \end{tabular}
\end{table}

\subsection{Office31: Amazon, Webcam and DSLR}
\emph{Office31} is a popular benchmark for visual cross-domain adaptation, which contains 4652 images and 31 categories. \emph{Office31}  consists of three visual domains: (1) \emph{Amazon}, (2) \emph{Webcam} and (3) \emph{DSLR}, which are abbreviated as A, W and D, respectively. \emph{Amazon} images is downloaded from online merchants, \emph{Webcam} images are with low-resolution by a web camera, and \emph{DSLR} images are with high-resolution by a digital SLR camera. We only focus on four challenging unsupervised domain adaption tasks among the three domains: A$\rightarrow$W, W$\rightarrow$A, A$\rightarrow$D and D$\rightarrow$A.
To alleviate the effect of the ill-generated images on classification, we train a classifier on the feature map learned by the generator, as in ADGAN \cite{sankaranarayanan2018generate}. For good performance, we initialize the generators $G_s,G_t$ by using the pre-trained ResNet-50 model trained on Imagenet.

We compare our method with seven latest and most-related methods, DANN \cite{ganin2016domain}, ADDA \cite{tzeng2017adversarial}, JAN \cite{long2017deep}, CDAN \cite{long2018conditional}, ADGAN \cite{sankaranarayanan2018generate}, and JDDA \cite{chen2019joint}. For fair comparison, we apply the standard protocol for unsupervised domain adaptation same as the baselines, and use all labeled source data and all unlabeled target data for experiments. We evaluate the averaged classification results of our method on three independent runs and report the best results of the baselines in the literature in Table \ref{officerst}. As observed, our method performs best in the two tasks (W$\rightarrow$A and W$\rightarrow$D), and is comparable with CDAN in the averaged results.

\subsection{More Results}
\subsubsection{\textbf{Ablation Study}}

To better demonstrate the performance of our method, we investigate different loss terms of our method in Eq. (\ref{objective}), and report the classification accuracy of their nine different combinations on \emph{Digits}, \emph{Faces} and \emph{Scenes} in Tables \ref{ablationstudy1} and \ref{ablationstudy2}, respectively. We observe that the more terms are added, the better the performance is. Also, our method (Row 9) shows the best performance among all the methods.
\begin{enumerate}[(1)]
\item Specifically, the first three rows show that bi-directional cycle-consistent GAN outperforms the single source-to-target or target-to-source ones;
\item Rows 4-6 show that the addition of the \emph{CLC} loss and \emph{JAG} loss brings a large performance improvement, respectively;
\item Rows 7-9 show that using an entropy-based confidence weight ($\gamma\leq 1$) to measure the reliability of predicting target samples is useful to align the joint distributions, and the \emph{DSP} loss also presents a significant effect to enhance the performance of our method;
\item In addition, comparing all these methods, we notice that introducing $\mathcal{L}_{\text{JGAN}_{t/s}}$ (comparing Rows 3 and 5) into our method achieves the most significant performance improvement in all terms, which implies that applying \emph{JAG} loss for the joint distribution alignment is most effective to reduce the classification deviation caused by domain gap.
\end{enumerate}

In sum, the results of Tables \ref{ablationstudy1} and \ref{ablationstudy2} show that all the proposed loss terms are useful, and they collectively form an effective solution of UDA.

\subsubsection{\textbf{Distribution Discrepancy}}

\begin{figure*}
    \centering
    \includegraphics[width=1.6\columnwidth]{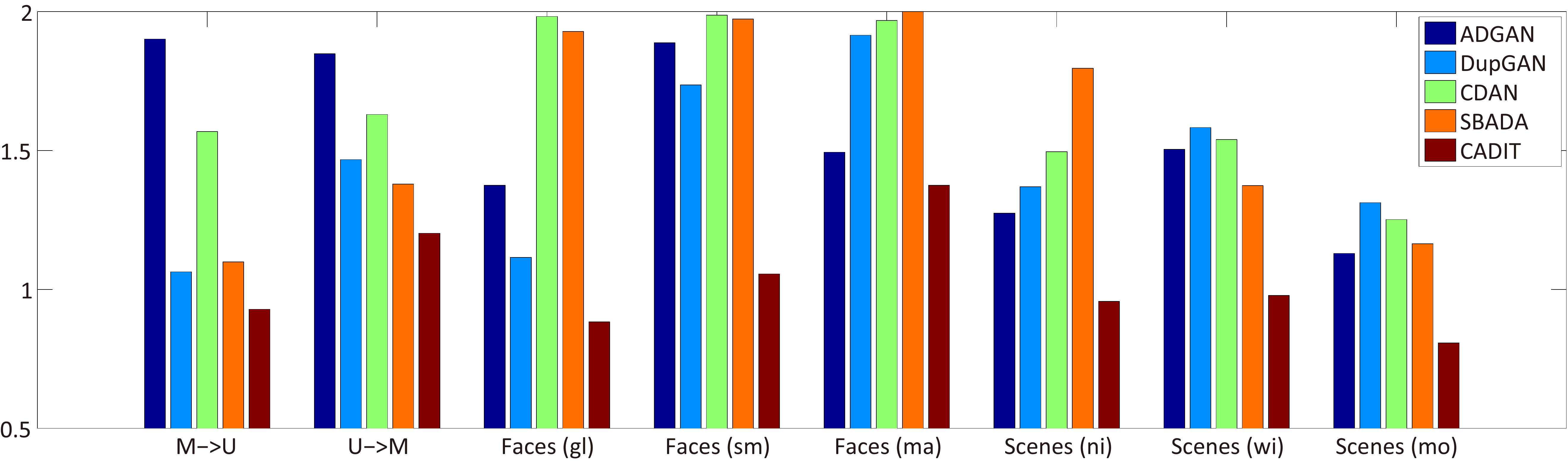}
    \caption{\small{The proxy $\mathcal{A}$-distance (for distribution discrepancy between source and target domains) on the eight tasks: (1)-(2) digital image recognition from MNIST to USPS and USPS to MNIST, (3)-(5) three classification tasks `glasses', `smiling' and `male' of facial images from photo to sketch, and (6)-(8) three classification tasks `night', `winter' and `mountain' of scene images from photo to painting, respectively.}}\label{distributiondiscrepancy}
\end{figure*}

To better show the performance of our method for reducing domain gap, we investigate the distribution discrepancy of domain-adaptated samples of our method on \emph{Digits}, \emph{Faces} and \emph{Scenes} data, compared to the related methods ADGAN, DupGAN, CDAN and SBADA. For SBADA and our CADIT, we extract the features from the images translated from source-to-target; for ADGAN, DupGAN and CDAN, we utilize the intermediate results generated by the domain adaptation models as their latent feature representations, due to their possible inferior/absent generated images. According to the previous theoretical analysis of domain adaptation \cite{ben2010theory,ganin2016domain}, the proxy $\mathcal{A}$-distance is adopted in our experiments, and the lower the $\mathcal{A}$-distance, the less the domain discrepancy. The $\mathcal{A}$-distance can be calculated as $d_{\mathcal{A}}=2(1-2\epsilon)$, where $\epsilon$ is the generalization error of the domain classifier \cite{ganin2016domain}. Fig. \ref{distributiondiscrepancy} depicts the $\mathcal{A}$-distance of all the five methods. As observed, $d_{\mathcal{A}}$ in our CADIT is lower than those in other four methods for all the tasks, which again indicates that our method can effectively reduce the domain discrepancy.

\begin{table*}
 \renewcommand{\arraystretch}{1}
 \setlength{\tabcolsep}{13pt}
\begin{center}
\caption{The classification accuracy (mean$\pm$std\%) in \emph{Office31} dataset. The top two results are in bold and underline, respectively.}\label{officerst}
\begin{tabular}{lccccc}\hline
\textbf{Methods}&\textbf{A$\rightarrow$W}&\textbf{W$\rightarrow$A}&\textbf{A$\rightarrow$D}&\textbf{D$\rightarrow$A}&\textbf{Average}\\\hline\hline
\textbf{DANN} (2016)&82.5$\pm$0.4&67.4$\pm$0.5&79.7$\pm$0.4&68.2$\pm$0.4&74.45\\\hline
\textbf{ADDA} (2017)&86.2$\pm$0.5&68.9$\pm$0.5&77.8$\pm$0.3&69.5$\pm$0.4&75.60\\\hline
\textbf{JAN} (2017)&85.4$\pm$0.3&70.0$\pm$0.4&84.7$\pm$0.3&68.6$\pm$0.3&77.18\\\hline
\textbf{CDAN} (2018)&\bm{94.1}$\pm$0.1&69.3$\pm$0.3&\bm{92.9}$\pm$0.2&71.0$\pm$0.3&\bm{81.83}\\\hline
\textbf{ADGAN} (2018)&89.5$\pm$0.5&\underline{71.4}$\pm$0.4&87.7$\pm$0.5&\underline{72.8}$\pm$0.3&80.35\\\hline
\textbf{JDDA} (2019)&82.6$\pm$0.4&66.7$\pm$0.2&79.8$\pm$0.1&57.4$\pm$0.0&71.63\\\hline
\textbf{CADIT}&\underline{90.4}$\pm$0.4&\bm{72.1}$\pm$0.3&\underline{89.2}$\pm$0.6&\bm{73.5}$\pm$0.2&\underline{81.31}\\\hline
\end{tabular}
\end{center}
\end{table*}

\begin{table*}
\begin{center}
\caption{\small{The results of the ablation study of our method: the classification accuracy of nine different combinations of the loss terms on \emph{Digits}. The results are in percent, and the best results are in bold.}}\label{ablationstudy1}
\renewcommand{\arraystretch}{1}
\setlength{\tabcolsep}{6pt}
\begin{tabular}{cccccccccccccc}\hline
\multirow{2}{*}{\textbf{}}&\multirow{2}{*}{\textbf{$\mathcal{L}_{\text{DGAN}_t}$}}&\multirow{2}{*}{\textbf{$\mathcal{L}_{\text{DGAN}_s}$}}&\multirow{2}{*}{\textbf{$\mathcal{L}_{\text{cyc}}$}}&\multirow{2}{*}{\textbf{$\mathcal{L}_{\text{CLC}}$}}&\multicolumn{2}{c}{\textbf{$\mathcal{L}_{\text{JGAN}_{s/t}}$}}&\multirow{2}{*}{\textbf{$\mathcal{L}_{\text{DSP}}$}}&&\multirow{2}{*}{\textbf{M$\rightarrow$U}}&\multirow{2}{*}{\textbf{U$\rightarrow$M}}&&\multirow{2}{*}{\textbf{S$\rightarrow$M}}&\multirow{2}{*}{\textbf{M$\rightarrow$S}}\\\cline{6-7}
&&&&&\textbf{$\gamma=1$}&\textbf{$\gamma\leq1$}&&&&&&&\\\hline\hline
1&\checkmark&&&&&&&&88.22&86.10&&67.17&35.73\\
2&&\checkmark&&&&&&&85.83&82.20&&53.29&34.67\\
3&\checkmark&\checkmark&\checkmark&&&&&&90.06&88.70&&75.07&50.18\\
4&\checkmark&\checkmark&\checkmark&\checkmark&&&&&93.83&91.10&&77.27&52.80\\
5&\checkmark&\checkmark&\checkmark&&\checkmark&&&&94.06&92.20&&81.36&55.98\\
6&\checkmark&\checkmark&\checkmark&\checkmark&\checkmark&&&&96.11&94.40&&83.08&57.02\\
7&\checkmark&\checkmark&\checkmark&\checkmark&&\checkmark&&&96.36&94.57&&87.03&59.08\\
8&\checkmark&\checkmark&\checkmark&\checkmark&\checkmark&&\checkmark&&96.48&94.81&&88.96&60.11\\
9&\checkmark&\checkmark&\checkmark&\checkmark&&\checkmark&\checkmark&&\textbf{96.74}&\textbf{95.04}&&\textbf{92.61}&\textbf{61.47}\\\hline
\end{tabular}
\end{center}
\end{table*}

\begin{table*}
\begin{center}
\caption{\small{The results of the ablation study of our method: the classification accuracy of nine different combinations of the loss terms on \emph{Faces} and \emph{Scenes}. The results are in percent, and the best results are in bold.}}\label{ablationstudy2}
\renewcommand{\arraystretch}{1}
\setlength{\tabcolsep}{5.5pt}
\begin{tabular}{ccccccccccccccc}\hline
\multirow{2}{*}{\textbf{}}&\multirow{2}{*}{\textbf{$\mathcal{L}_{\text{DGAN}_t}$}}&\multirow{2}{*}{\textbf{$\mathcal{L}_{\text{DGAN}_s}$}}&\multirow{2}{*}{\textbf{$\mathcal{L}_{\text{cyc}}$}}&\multirow{2}{*}{\textbf{$\mathcal{L}_{\text{CLC}}$}}&\multicolumn{2}{c}{\textbf{$\mathcal{L}_{\text{JGAN}_{s/t}}$}}&\multirow{2}{*}{\textbf{$\mathcal{L}_{\text{DSP}}$}}&\multicolumn{3}{c}{\textbf{\emph{Faces}}}&&\multicolumn{3}{c}{\textbf{\emph{Scenes}}}\\\cline{6-7}\cline{9-11}\cline{13-15}
&&&&&\textbf{$\gamma=1$}&\textbf{$\gamma\leq1$}&&\textbf{glasses}&\textbf{smiling}&\textbf{male}&&\textbf{night}&\textbf{winter}&\textbf{mountain}\\\hline\hline
1&\checkmark&&&&&&&87.28&74.31&70.56&&74.68&62.64&66.96\\
2&&\checkmark&&&&&&86.23&72.12&68.28&&71.94&60.76&65.87\\
3&\checkmark&\checkmark&\checkmark&&&&&89.95&75.46&71.97&&76.76&64.55&68.10\\
4&\checkmark&\checkmark&\checkmark&\checkmark&&&&90.69&76.23&72.88&&78.43&65.45&69.01\\
5&\checkmark&\checkmark&\checkmark&&\checkmark&&&92.26&78.35&74.51&&80.89&67.31&71.06\\
6&\checkmark&\checkmark&\checkmark&\checkmark&\checkmark&&&94.20&82.30&76.08&&83.02&70.10&73.03\\
7&\checkmark&\checkmark&\checkmark&\checkmark&&\checkmark&&95.22&84.12&78.89&&84.05&71.04&74.01\\
8&\checkmark&\checkmark&\checkmark&\checkmark&\checkmark&&\checkmark&96.02&85.05&79.84&&85.09&72.01&75.10\\
9&\checkmark&\checkmark&\checkmark&\checkmark&&\checkmark&\checkmark&\textbf{97.11}&\textbf{86.16}&\textbf{80.91}&&\textbf{86.05}&\textbf{73.14}&\textbf{76.05}\\\hline
\end{tabular}
\end{center}
\end{table*}
\section{Conclusion}\label{conclusions}
We present a novel method to simultaneously align the joint and marginal distributions between domains for adversarial image translation in order to better match the class distributions that is important for classification. Specifically, we develop the \emph{JAG} loss, \emph{DSP} loss and \emph{CLC} loss on top of the structure of CycleGAN. Extensive experiments validate that our method achieves the outstanding performance in \emph{Faces} and \emph{Scenes} and the competitive results in \emph{Digits} and \emph{Office31}, compared with the state-of-the-arts. Also, the ablation study verifies that the proposed \emph{JAG} loss (for the joint distribution alignment) in our method is indeed effective to enhance the classification accuracy, and the proposed \emph{DSP} loss is also useful to prevent label flipping.


\ifCLASSOPTIONcaptionsoff
  \newpage
\fi

\bibliographystyle{IEEEtran}
\bibliography{IEEEabrv,egbib}

\end{document}